\documentclass[10pt,conference,english]{IEEEtran}
\usepackage[utf8]{inputenc}
\usepackage[english]{babel}
\usepackage[T1]{fontenc}
\usepackage{amsfonts}
\usepackage{amsmath}
\usepackage{amssymb}
\usepackage{tikz}
\usepackage{graphicx}
\usepackage{color}
\usepackage{pgfpages}
\usepackage[keeplastbox]{flushend}
\usetikzlibrary{arrows}
\newcommand\Mark[1]{\textsuperscript{#1}}

\begin{document}

\title{Learning Local Receptive Fields and their\\Weight Sharing Scheme on Graphs\\[.75ex]
  {\normalfont\large 
    Jean-Charles Vialatte\Mark{1,2}, Vincent Gripon\Mark{2}, Gilles Coppin\Mark{2}%
  }\\[-1.5ex]
}

\author{
    \IEEEauthorblockA{%
        \Mark{1}City Zen Data\\
        55 rue Charles Nungesser\\
        29490 Guipavas, France\\
        \texttt{jean-charles.vialatte@cityzendata.com}
    }
    \and
    \IEEEauthorblockA{%
      \Mark{2}IMT Atlantique / CNRS Lab-STICC\\
      Technopole Brest Iroise\\
      29238 Brest, France\\
      \texttt{name.surname@imt-atlantique.fr}
    }
}

\maketitle

\begin{abstract}
   We propose a simple and generic layer formulation that extends the properties of convolutional layers to any domain that can be described by a graph. Namely, we use the support of its adjacency matrix to design learnable weight sharing filters able to exploit the underlying structure of signals in the same fashion as for images. The proposed formulation makes it possible to learn the weights of the filter as well as a scheme that controls how they are shared across the graph. We perform validation experiments with image datasets and show that these filters offer performances comparable with convolutional ones.
\end{abstract}


\section{Introduction}
Convolutional Neural Networks (CNNs) have achieved state-of-the-art accuracy in many supervised learning challenges~\cite{lecun2010convolutional,cirecsan2011committee,krizhevsky2012imagenet,sainath2013deep,farabet2013learning,tompson2014joint}. For their ability to absorb huge amounts of data with lesser overfitting, deep learning~\cite{lecun2015deep} models are the golden standard when a lot of data is available. CNNs benefit from the ability to create stationary and multi-resolution low-level features from raw data, independently from their location in the training images. Some authors draw a parallel between these features and scattering transforms~\cite{mallat2016understanding}.

Obviously CNNs rely on the ability to define a convolution operator (or a translation) on signals. On images, this amounts to learn local receptive fields~\cite{moody1988learning} that are convolved with training images. Considering images to be defined on a grid graph, we point out that the receptive fields of vertices are included in their neighbors -- or, more generally, a neighborhood.

Reciprocally, convolution requires more than the neighborhoods of vertices in the underlying graph, as the operator is able to match specific neighbors of distinct vertices together. For instance, performing convolution on images requires the knowledge of coordinates of pixels, that is not directly accessible when considering a grid graph (c.f.~\cite{GrePasViaGri201610,vialatte2016generalizing}). In this paper we are interested in demonstrating that the underlying graph is nevertheless enough to achieve comparable results.

The convolution of a signal can be formalized as its multiplication with a convolution matrix. In the case of images and for small convolution kernels, it is interesting to note that this convolution matrix has the same support as a lattice graph. Using this idea, we propose to introduce a type of layer based on a graph that connects neurons to their neighbors. Moreover, convolution matrices are entirely determined by a single row, since the same weights appear on each one. To imitate this process, we introduce a weight sharing learning procedure, that consists in using a limited pool of weights that each row of the obtained operator can make use of.

Section II presents related work. Section III describes our methodology and the links with existing architectures. Section IV contains experimental results. Section V is a conclusion.

\section{Related Work}

Due to the effectiveness of CNNs on image datasets, models have been proposed to adapt them to other kind of data, e.g. for shapes and manifolds \cite{masci2015shapenet,monti2016geometric}, molecular datasets \cite{duvenaud2015convolutional}, or graphs \cite{niepert2016learning,atwood2016diffusion,monti2016geometric}. 
A review is done in~\cite{bronstein2016geometric}. 
In particular, CNNs have also been adapted to graph signals, such as in \cite{bruna2013spectral,henaff2015deep} where the convolution is formalized in the spectral domain of the graph defined by its Laplacian \cite{shuman2013emerging}.
This approach have been improved in \cite{defferrard2016convolutional}, with a localized and fast approximated formulation, and has been used back in vision to breed isometry invariant representations~\cite{khasanova2017graph}.

For non-spectral approaches, feature correspondences in the input domain allow to define how the weights are tied across the layer, such as for images or manifolds. For graphs and graph signals, such correspondences doesn't necessarily exist. For example, in \cite{atwood2016diffusion} (where the convolution is based on multiplications with powers of the probability transition matrix) weights are tied according to the power to which they are attached, in \cite{niepert2016learning} an ordering of the nodes is used, in \cite{monti2016geometric} an embedding is learned from the degrees of the nodes. These choices are arbitrary and unsimilar to what is done by regular convolutions. On the contrary, we propose a generic layer formulation that allows to also learn how the weights are linearly distributed over the local receptive field.

Our model is first designed for the task of graph signal classification, but another common task is the problem of node classification such as in \cite{atwood2016diffusion,yang2016revisiting,kipf2016semi,monti2016geometric}. Models learning part of their structures have also been proposed, such as in \cite{feng2015learning,cortes2016adanet}. Moreover, because our model strongly ressembles regular convolutions, it can also ressemble some of their variants, such as group equivariant convolutions~\cite{cohen2016group}.

\section{Methodology}

We first recall the basic principles of Deep Neural Networks (DNNs) and CNNs, then introduce our proposed graph layer.

\subsection{Background}
DNNs~\cite{hornik1989multilayer} consist of a composition of layers, each one parametrized by a learnable weight kernel $W$ and a nonlinear function $f: \mathbb{R}\to\mathbb{R}$. Providing the input of such a layer is $\mathbf{x}$, the corresponding output is then:
\[
\mathbf{y} = f(W \cdot \mathbf{x} + \mathbf{b})\;,
\]
where $\cdot$ is the matrix product operator, $f$ is typically applied component-wise and $\mathbf{b}$ is a learnable bias vector.

The weight kernels are learned using an optimization routine usually based on gradient descent, so that the DNN is able to approximate an objective function. A DNN containing only this type of layer is called Multi-Layer Perceptron (MLP).

\begin{figure}[h]
  \begin{center}
    \begin{tikzpicture}
      \tikzstyle{every node} = [draw, circle, thick, inner sep = 2pt]
      \foreach \y in {0,...,4}{
        \pgfmathtruncatemacro{\yplusone}{5 - \y}
        \node(a\y) at (0,.6*\y) {\footnotesize\yplusone};
      }
      \foreach \y in {0,...,4}{
        \pgfmathtruncatemacro{\yplusone}{5 - \y}
        \node(\y) at (2,.6*\y) {\footnotesize\yplusone};
      }
      \path[opacity=0.5]
      (a0) edge (0);
      \path[dashed]
      (a0) edge (1);
      \path[dotted]
      (a1) edge (0);
      \path[opacity=0.5]
      (a1) edge (1);
      \path[dashed]
      (a1) edge (2);
      \path[dotted]
      (a2) edge (1);
      \path[opacity=0.5]
      (a2) edge (2);
      \path[dashed]
      (a2) edge (3);
      \path[dotted]
      (a3) edge (2);
      \path[opacity=0.5]
      (a3) edge (3);
      \path[dashed]
      (a3) edge (4);
      \path[dotted]
      (a4) edge (3);
      \path[opacity=0.5]
      (a4) edge (4);
      \tikzstyle{every node} = []
      \node at (5,1.2) {$\left(\begin{array}{ccccc}
          w_{2} & w_{3} & 0 & 0 & 0\\
          w_1 & w_{2} & w_{3} & 0 & 0\\
          0 & w_1 & w_{2} & w_{3} & 0\\
          0 & 0 & w_1 & w_{2} & w_{3}\\
          0 & 0 & 0 & w_{1} & w_{2}
        \end{array}\right)$};
    \end{tikzpicture}
  \end{center}
  \caption{Depiction of a 1D-convolutional layer and its associated matrix $W$.}
  \label{cnn}
\end{figure}
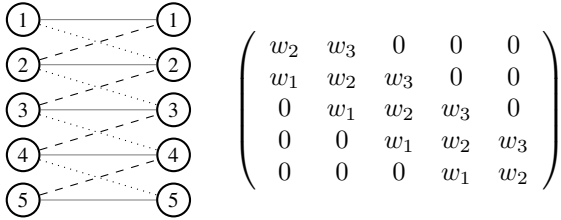

In the case of CNNs~\cite{lecun1998gradient}, some of the layers have the particular form of convolution filters. In this case, the convolutional operation can also be written as the product of the input signal with a matrix $W$, where $W$ is a Toeplitz matrix. Previous works~\cite{graham2014spatially,simonyan2014very,springenberg2014striving,he2016deep} have shown that to obtain the best accuracy in vision challenges, it is usually better to use very small kernels, resulting in a sparse $W$. Figure~\ref{cnn} depicts a convolutional layer.

\subsection{Proposed Method}

We propose to introduce another type of layer, that we call \textit{receptive graph layer}. It is based on an adjacency matrix and aims at extending the principle of convolutional layers to any domain that can be described using a graph.

Consider an adjacency matrix $A$ that is well fitted to the signals to be learned, in the sense that it describes an underlying graph structure between the input features. We define the receptive graph layer associated with $A$ using the product between a third rank tensor $S$ and a weight kernel $W$. For now, the tensor $W$ would be one-rank containing the weights of the layer and $S$ is of shape $n\times n \times \omega$, where $n\times n$ is the shape of the adjacency matrix and $\omega$ is the shape of $W$.

On the first two ranks, the support of $S$ must not exceed that of $A$, such that $A_{ij} = 0 \Rightarrow \forall k, S_{ijk} = 0$.

Overall, we obtain:
\[
\mathbf{y} = f(W\cdot S \cdot \mathbf{x} + \mathbf{b})\;,
\]
where here $\cdot$ denotes the tensor product.

\begin{figure}[h]
  \begin{center}
    \begin{tikzpicture}
      \tikzstyle{every node} = [draw, circle, thick, inner sep = 2pt]
      \begin{scope}[yshift=3.5cm,xshift=-3cm]
        \foreach \y in {0,...,4}{
          \pgfmathtruncatemacro{\yplusone}{\y + 1}
          
          \node(\y) at (.8*\y,0) {\footnotesize\yplusone};
        }
        \path
        (0) edge[loop above] (0)
        edge (1);
        \path
        (1) edge[loop above] (1)
        edge (2);
        \path
        (2) edge[loop above] (2)
        edge (3);
        \path
        (3) edge[loop above] (3)
        edge (4);
        \path
        (4) edge[loop above] (4);
      \end{scope}
      \begin{scope}[yshift=2.5cm, xshift=1.5cm]
        \foreach \y in {0,...,4}{
          \pgfmathtruncatemacro{\yplusone}{5 - \y}

          \node(a\y) at (0,.5*\y) {\footnotesize\yplusone};
      }
        \foreach \y in {0,...,4}{
          \pgfmathtruncatemacro{\yplusone}{5 - \y}
                  
          \node(\y) at (2,.5*\y) {\footnotesize\yplusone};
      }
      \path
      (a0) edge (0)
      (a0) edge (1);
      \path
      (a1) edge (0)
      (a1) edge (1)
      (a1) edge (2);
      \path
      (a2) edge (1)
      (a2) edge (2)
      (a2) edge (3);
      \path
      (a3) edge (2)
      (a3) edge (3)
      (a3) edge (4);
      \path
      (a4) edge (3)
      (a4) edge (4);
      \end{scope}
      \tikzstyle{every node} = []
      \node at (-1,1) {$\left(\begin{array}{ccccc}
          \mathbf{s}_{11} & \mathbf{s}_{12} & 0 & 0 & 0\\
          \mathbf{s}_{21} & \mathbf{s}_{22} & \mathbf{s}_{23} & 0 & 0\\
          0 & \mathbf{s}_{32} & \mathbf{s}_{33} & \mathbf{s}_{34} & 0\\
          0 & 0 & \mathbf{s}_{43} & \mathbf{s}_{44} & \mathbf{s}_{45}\\
          0 & 0 & 0 & \mathbf{s}_{54} & \mathbf{s}_{55}
        \end{array}\right)$};
      \node(label) at (3.2,1.5) {$(W\cdot S)_{45} = \displaystyle{\sum_{k=1}^{\omega}{W_{k} S_{45 k}}}$};
      \path[>=stealth, ->, thick]
      (label) edge[bend left] (2.9,2.65);
    \end{tikzpicture}
  \end{center}
  \caption{Depiction of a graph, the corresponding receptive graph of the propagation and its associated weight sharing scheme $S$. Note that $\mathbf{s}_{ij}=S_{ij\cdot}$ are vector slices of $S$ along the first two ranks, $\mathbf{s}_{ij}$ determines how much of each weight in $W$ is allocated for the edge linking vertex $i$ to vertex $j$.}
  \label{proposed}
\end{figure}
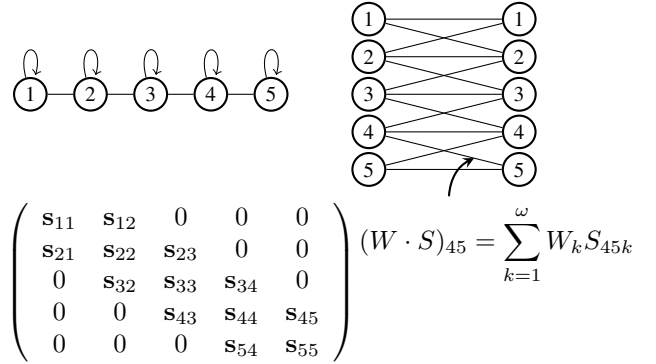

Intuitively, the values of the weight kernel $W$ are linearly distributed to pairs of neighbours in $A$ with respect to the values of $S$. For this reason, we call $S$ the \textit{scheme} (or \textit{weight sharing scheme}) of the receptive graph. In a sense, this scheme tensor is to the receptive graph what the adjacency matrix is to the graph. An example is depicted in Figure~\ref{proposed}.

Alike convolution on images, $W$ is extended as a third-rank tensor to include multiple input and output channels (also known as feature maps). It is worth mentioning that an implementation must be memory efficient to take care of a possibly large sparse $S$.

\subsection{Training}

The proposed formulation allows to learn both $S$ and $W$. We perform the two jointly. Learning $W$ amounts to learning weights as in regular CNNs, whereas learning $S$ amounts to learning how these weights are tied over the receptive fields. We also experiment a fine-tuning step, which consists in freezing $S$ in the last epochs. Indeed, when a weight sharing scheme can be decided directly from the underlying structure, it is not necessary to train $S$.

Because of our inspiration from CNNs, we propose constraints on the parameters of $S$. Namely, we impose them to be between 0 and 1, and to sum to 1 along the third dimension. Therefore, the vectors on the third rank of $S$ can be interpreted as performing a weighted average of the parameters in $W$.

We test two types of initialization for $S$. The first one consists in distributing one-hot-bit vectors along the third rank. We impose that for each receptive field, a particular one-hot-bit vector can only be distributed at most once more than any other. We refer to it as one-hot-bit initialization. The second one consists in using a uniform random distribution with limits as described in \cite{glorot2010understanding}.

\subsection{Genericity}

For simplicity we restricted our explanation to square adjacency matrices. In the case of oriented graphs, one could remove the rows and columns of zeros and obtain a receptive graph with a distinct number of neurons in the input ($n$) than in the output ($m$). As a result, receptive graph layers extend usual ones, as explained here:
\begin{enumerate}
\item To obtain a fully connected layer, one can choose $\omega$ to be of size $nm$ and $S$ the matrix of vectors that contains all possible one-hot-bit vectors.
\item To obtain a convolutional layer, one can choose $\omega$ to be the size of the kernel. $S$ would be one-hot-bit encoded along its third rank and circulant along the first two ranks. A stride $> 1$ can be obtained by removing the corresponding rows.
\item Similarly, most of the layers presented in related works can be obtained for an appropriate definition of $S$.
\end{enumerate}

In our case, $S$ is more similar to that obtained when considering convolutional layers, with the noticeable differences that we do not force which weight to allocate for which neighbor along its third rank and it is not necessarily circulant along the first two ranks.

\subsection{Discussion}
\label{discussion}

Although we train $S$ and $W$, the layer propagation is ultimately handled by their tensor product. That is, its output is determined by $\Theta \cdot \mathbf{x}$ where $\Theta = S \cdot W$. For the weight sharing to make sense, we must then not over-parameterize $S$ and $W$ over $\Theta$. If we call $l$ the number of non-zeros in $A$ and $w\times p \times q$ the shape of $W$, then the former assumption requires $lw + wpq \leq lpq$ or equivalently $\frac{1}{w} \geq \frac{1}{pq} + \frac{1}{l}$. It implies that the number of weights per filter $w$ must be lower than the total number of filters $pq$ and than the number of edges $l$.

Note that without the constraint that the support of $S$ must not exceed that of $A$ (or if the used graph is complete), the proposed formulation could also be applied to structure learning of the input features space~\cite{richardson1996discovery,kwok1997constructive}. That is, operations on $S$ along the third rank might be exploitable in some way, e.g. dropping connections during training~\cite{han2015learning} or discovering some sort of structural correlations. However, even if this can be done for toy image datasets, such $S$ wouldn't be sparse and would lead to memory issues in higher dimensions. So we didn't include these avenues in the scope of this paper.

\section{Experiments}

\subsection{Description}

We are interested in comparing various receptive graph layers with convolutional ones. For this purpose, we use image datasets, but restrain priors about the underlying structure.

We first present experiments on MNIST~\cite{lecun1998mnist}. It contains 10 classes of gray levels images (28x28 pixels) with 60'000 examples for training, 10'000 for testing. We also do experiments on a scrambled version to hide the underlying structure, as done in previous work~\cite{chen2014unsupervised}. Then we present experiments on Cifar10~\cite{krizhevsky2009learning}. It contains 10 classes of RGB images (32x32 pixels) with 50'000 examples for training, 10'000 for testing.

Because receptive graph layers are wider than their convolutional counterparts ($lw$ more parameters from $S$), experiments are done on shallow (but wide) networks for this introductory paper. Also note that they require $w+1$ times more multiply operations than a convolution lowered to a matrix multiplication~\cite{chetlur2014cudnn}. In practice, they roughly took 2 to 2.5 more time.

\subsection{Experiments with grid graphs on MNIST}

Here we use models composed of a single receptive graph (or convolutional) layer made of 50 feature maps, without pooling, followed by a fully connected layer of 300 neurons, and terminated by a softmax layer of 10 neurons. Rectified Linear Units~\cite{glorot2011deep} are used for the activations and a dropout~\cite{srivastava2014dropout} of 0.5 is applied on the fully-connected layer. Input layers are regularized by a factor weight of $10^{-5}$~\cite{ng2004feature}. We optimize with ADAM~\cite{kingma2014adam} up to 100 epochs and fine-tune (while $S$ is frozen) for up to 50 additional epochs.

We consider a grid graph that connects each pixel to itself and its 4 nearest neighbors (or less on the borders). We also use the square of this graph (pixels are connected to their 13 nearest neighbors, including themselves), the cube of this graph (25 nearest neighbors), up to 10 powers (211 nearest neighbors).
Here we use one-hot-bit initialization. We test the model under two setups: either the ordering of the node is unknown, and then the one-hot-bit vectors are distributed randomly and modified upon training ; either an ordering of the node is known, and then the one-hot-bit vectors are distributed in a circulant fashion in the third rank of $S$ which is freezed in this state. We use the number of nearest neighbors as for the dimension of the third rank of $S$.
We also compare with a convolutional layer of size 5x5, thus containing as many weights as the cube of the grid graph. Table~\ref{toy} summarizes the obtained results. The ordering is unknown for the first result given, and known for the second result between parenthesis.

\begin{table}[h]
  \caption{Error rates on powers of the grid graphs on MNIST.}
  \begin{center}
    \bgroup
    \def\arraystretch{1.5}
    \begin{tabular}{|c|c|c|c|}
      \hline
      Conv5x5 & Grid$^1$ & Grid$^2$ & Grid$^3$\\
      \hline
      (0.87\%) & 1.24\% (1.21\%) & 1.02\% (0.91\%) & 0.93\% (0.91\%)\\
      \hline
      \hline
      Grid$^4$ & Grid$^5$ & Grid$^6$ & Grid$^{10}$\\
      \hline
      0.90\% (0.87\%) & 0.93\% (0.80\%) & 1.00\% (0.74\%) & 0.93\% (0.84\%)\\
      \hline
    \end{tabular}
    \egroup
  \end{center}
  \label{toy}
\end{table}

We observe that even without knowledge of the underlying euclidean structure, receptive grid graph layers obtain comparable performances as convolutional ones, and when the ordering is known, they match convolutions. We also noticed that after training, even though the one-hot-bit vectors used for initialization had changed  to floating point values, their most significant dimension was always the same. That suggests there is room to improve the initialization and the optimization.

\begin{figure}[h]
  \begin{center}
    \input{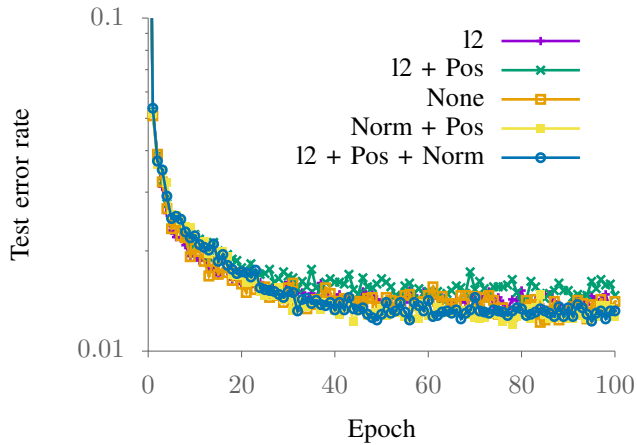}
  \end{center}
  \caption{Evolution of the test error rate when learning MNIST using the square of a grid graph and for various normalizations, as a function of the epoch of training. The legend reads: ``l2'' means $\ell_2$ normalization of weights is used (with weights $10^{-5}$), ``Pos'' means parameters in $S$ are forced to being positive, and ``Norm'' means that the $\ell_1$ norm of each vector in the third dimension of $S$ is forced to 1.}
  \label{functionofepoch}
\end{figure}

In Figure~\ref{functionofepoch}, we plot the test error rate for various normalizations when using the square of the grid graph, as a function of the number of epochs of training. We observe that they have little influence on the performance and sometimes improve it a bit. Thus, we use them as optional hyperparameters.

\subsection{Experiments with covariance graphs on Scrambled MNIST}

 We use a thresholded covariance matrix obtained by using all the training examples. We choose the threshold so that the number of remaining edges corresponds to a certain density $p$ (5x5 convolutions correspond approximately to a density of $p=3\%$). We also infer a graph based on the $k$ nearest neighbors of the inverse of the values of this covariance matrix ($k$-NN). The latter two are using no prior about the signal underlying structure. The pixels of the input images are shuffled and the same re-ordering of the pixels is used for every image. Dimension of the third rank of $S$ is chosen equal to $k$ and its weights are initialized random uniformly~\cite{glorot2010understanding}.
 The receptive graph layers are also compared with models obtained when replacing the first layer by a fully connected or convolutional one. Architecture used is the same as in the previous section. Results are reported on table~\ref{covar}.

\begin{table}[h]
  \caption{Error rates when topology is unknown on scrambled MNIST.}
  \begin{center}
    \bgroup
    \def\arraystretch{1.5}
    \begin{tabular}{|c|c|c|c|}
      \hline
      MLP & Conv5x5 & Thresholded ($p=3\%$) & $k$-NN ($k=25$)\\
      \hline
      1.44\% & 1.39\% & 1.06\% & 0.96\%\\
      \hline
    \end{tabular}
    \egroup
  \end{center}
  \label{covar}
  \end{table}

We observe that the receptive graph layers outperforms the CNN and the MLP on scrambled MNIST. This is remarkable because that suggests it has been able to exploit information about the underlying structure thanks to its graph.

\subsection{Experiments with shallow architectures on Cifar10}

On Cifar10, we made experiments on shallow CNN architectures and replaced convolutions by receptive graphs. We report results on a variant of AlexNet~\cite{krizhevsky2012imagenet} using little distortion on the input that we borrowed from a tutorial of tensorflow~\cite{tensorflow2015-whitepaper}.
It is composed of two 5x5 convolutional layers of 64 feature maps, with max pooling and local response normalization, followed by two fully connected layers of 384 and 192 neurons.
We compare two different graph supports: the one obtained by using the underlying graph of a regular 5x5 convolution, and the support of the square of the grid graph. Optimization is done with stochastic gradient descent on 375 epochs where $S$ is freezed on the 125 last ones. Circulant one-hot-bit intialization is used. These are weak classifiers for Cifar10 but they are enough to analyse the usefulness of the proposed layer. Exploring deeper architectures is left for further work. Experiments are run five times each. Means and standard deviations of accuracies are reported in table~\ref{cifar}. ``Pos'' means parameters in $S$ are forced to being positive, ``Norm'' means that the $\ell_1$ norm of each vector in the third dimension of $S$ is forced to 1, ``Both'' means both constraints are applied, and ``None'' means none are used.

\begin{table}[h]
  \caption{Accuracies (in \%) of shallow networks on CIFAR10.}
  \begin{center}
    \bgroup
    \def\arraystretch{1.5}
    \begin{tabular}{|c|c|c|c|c|c|c|}
      \hline
      Support & Learn $S$ & None & Pos & Norm & Both\\
      \hline
      \hline
      Conv5x5 & No & / & / & / & $86.8 \pm 0.2$\\
      \hline
      Conv5x5 & Yes & $87.4 \pm 0.1$ & $87.1 \pm 0.2$ & $87.1 \pm 0.2$ & $87.2 \pm 0.3$\\
      \hline
      Grid$^2$ & Yes & $87.3 \pm 0.2$ & $87.3 \pm 0.1$ & $87.5 \pm 0.1$ & $87.4 \pm 0.1$\\
      \hline
    \end{tabular}
    \egroup
  \end{center}
  \label{cifar}
\end{table}

The receptive graph layers are able to outperform the corresponding CNNs by a small amount in the tested configurations, opening the way for more complex architectures.

\section{Conclusion}

We introduced a new class of layers for deep neural networks which consists in using the support of a graph operator and linearly distributing a pool of weights over the defined edges. The linear distribution is learned jointly with the pool of weights. Thanks to these structural dependencies, we showed it is possible to share weights in a fashion similar to Convolutional Neural Networks (CNNs).

We performed experiments on vision datasets where the receptive graph layer obtains similar performance as convolutional ones, even when the underlying image structure is hidden. We believe that with further work, the proposed layer could fully extend the performance of CNNs to many other domains described by a graph.

Future works will also include exploration of more advanced graph inference techniques. One example is using gradient descent from the supervised task at hand~\cite{henaff2015deep}. We can also notice that in our case, this amounts to select receptive fields, breeding another avenue~\cite{coates2011selecting}.

\section*{Acknowledgments}
This work was funded in part by the CominLabs project Neural Communications, and by the ANRT (Agence Nationale de la Recherche et de la Technologie) through a CIFRE (Convention Industrielle de Formation par la REcherche).

\bibliographystyle{IEEEtran}
\bibliography{article}

\end{document}